# Towards a Multidimensional Evaluation Framework for Empathetic Conversational Systems

Aravind Sesagiri Raamkumar, and Siyuan Brandon Loh

Agency for Science, Technology and Research (A*STAR), Singapore

**Abstract.** Empathetic Conversational Systems (ECS) are built to respond empathetically to the user's emotions and sentiments, regardless of the application domain. Current ECS studies evaluation approaches are restricted to offline evaluation experiments primarily for gold standard comparison & benchmarking, and user evaluation studies for collecting human ratings on specific constructs. These methods are inadequate in measuring the actual quality of empathy in conversations. In this paper, we propose a multidimensional empathy evaluation framework with three new methods for measuring empathy at (i) structural level using three empathy-related dimensions, (ii) behavioral level using empathy behavioral types, and (iii) overall level using an empathy lexicon, thereby fortifying the evaluation process. Experiments were conducted with the state-of-the-art ECS models and large language models (LLMs) to show the framework's usefulness.

**Keywords:** Empathetic Conversational Systems, Empathetic Dialog Systems, Empathic Chatbot, Empathy Evaluation, Empathetic AI, Empathic AI

## 1    Introduction

Empathy is the ability to understand and share the feelings of another human. It involves being able to put oneself in someone else's shoes and perceive a situation from their perspective. Empathy enables humans to connect with others through understanding and experiencing their emotions. In summary, empathy allows humans to emotionally understand and relate to what someone else is going through or experiencing [1, 2]. While computational modeling of empathy aids in better understanding human relations [3], empathic ability has also been embedded in dialog/conversational systems. Empathetic conversational systems (ECS) or empathetic dialogue systems are conversational systems that are trained to respond empathetically to user utterances [4, 5]. A typical working representation of an ECS implementation is illustrated in Figure 1.

The most prevalent process flow in existing ECS studies is to train or fine-tune a pre-trained foundation model on datasets comprising empathetic conversations between two individuals and evaluate the model with traditional offline evaluation metrics followed by a high-level user evaluation study. This approach suffers from multiple issues. Firstly, the quality of empathy in the datasets is not explicitly earmarked, thereby the models trained on such datasets cannot differentiate a weak empathetic response from a strong one since the assumption is that all conversations in the datasets are highly empathetic. Secondly, the offline evaluation metrics predominantly compare the ECS-generated responses with gold-standard responses (human responses from the datasets) using metrics such as BLEU, and Perplexity, to name a few.

These metrics do not measure the level of empathy, but instead quantify the similarity between the ground truth and generated responses. Additionally, the user-evaluation studies elicit responses from participants for simple questions such as "Did the responses show understanding of the feelings of the person talking about their experience?". Hence, ECS models need to be evaluated in a comprehensive manner with empathy and its related concepts as the key focus area.

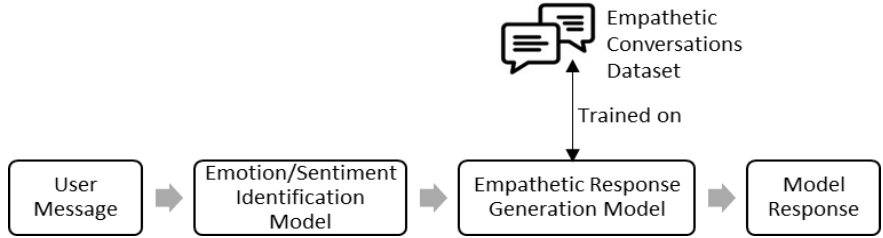

**Fig. 1.** Typical ECS Representation

In this paper, we aim to address the aforementioned issues by proposing a multidimensional empathy evaluation framework comprising five evaluation components. Out of these five evaluation components, three components are novel while the other two components are dedicated to the *traditional offline evaluation experiments* and *user studies* respectively. The three novel empathy-focused components measure empathy using (i) *structural empathy dimensions – emotional reaction, interpretation, and exploration, (ii) empathetic behavior types – empathetic concern, consolation, altruistic heling, perspective taking, mirroring, and (iii) empathy lexicon*. We have conducted experiments to highlight the differences between the components and the overall usefulness of the framework, for three of these components. We postulate that this framework will help researchers and practitioners in better evaluating empathetic responding capability of ECS models, and large language model (LLM) chatbots since it considers multiple aspects.

## 2 Multidimensional Empathy Evaluation Framework

The multidimensional empathy evaluation framework components are illustrated in Figure 2. The components are (i) structural empathy-based evaluation, (ii) empathetic behavior evaluation, (iii) empathy lexicon-based evaluation, (iv) offline evaluation, (v) human evaluation. The main goal of employing this framework in evaluation experiments is to ensure that future ECS studies yield models that perform well for the three empathy-related components. Additionally, if the human evaluation results are favorable, it can be concluded that the proposed models are suitable for the task of empathetic responding. The offline evaluation component becomes important if the gold-standard responses are measured to be highly empathetic. We conducted simple experiments for three components, by using the test set of the EmpatheticDialogues (ED) dataset [6]. The test set has around 5200 dialogs for 32 emotion classes. Each dialog comprises of seeker and supporter utterances in which the seeker was given one emotion class and requested to initiate a conversation pertaining to that particular emotion while the supporter was requested to be empathetic while responding.

We selected four SOTA ECS models:-

- The ED retrieval model [6]
- The Knowledge-aware EMPathetic dialogue generation method (KEMP) model [7]
- The Focused Empathy (FE) model [8]
- The Cognition, Affection and Behavior (CAB) model [9]

In addition, we selected LLMs GPT3.5 and Vicuna since LLMs can be directly used in zero-shot/few-shot setting to respond empathetically to users [10]. We also included a finetuned version of Vicuna which was finetuned on a sample set of conversations from the ED dataset.

- GPT3.5 (GPT),
- Vicuna FastChat-T5 (Vicuna)
- Vicuna (Vicuna_FT)

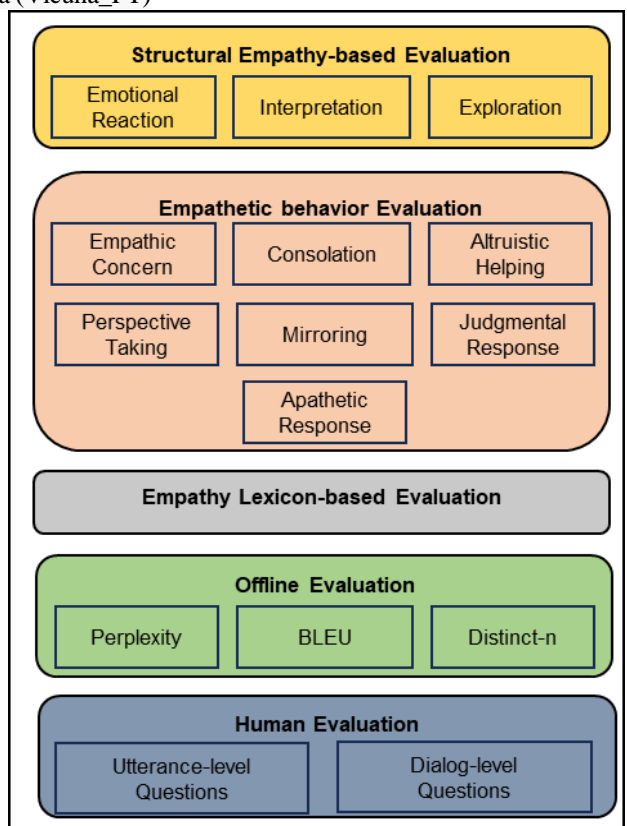

**Fig. 2.** Multidimensional Empathy Evaluation Framework

## 2.1 Structural Empathy-based Evaluation

There are different types or levels of empathy, namely affective empathy, cognitive empathy, and compassionate empathy [11]. It has already been highlighted that the current evaluation methods in ECS studies do not measure empathy at these levels. There is a necessity to consider at least affective and cognitive empathy for evaluation as these two types are important. There is an existing approach for measuring the empathy levels in natural language text using three empathy-related dimensions *emotional reaction* (proxy for affective empathy), *interpretation* (proxy for cognitive empathy), *and exploration* [12]. Emotional reaction measures the ability of the responder in addressing the emotional concerns of the seeker. Interpretation measures the responder's ability in reaffirming and understanding the issues of the seeker while exploration measures the ability to ask questions for deeper understanding. There are three output values for each dimension. They are strong, weak, and no signals.

These responses from models along with the gold-standard human responses from the ED dataset were compared using the three dimensions. The results are illustrated in Figure 3. From the chart, it can be deduced that the Vicuna LLMs produced better results for the two dimensions emotional reaction (*Vicuna_FT: 0.9, Vicuna: 0.81*) and interpretation (*Vicuna_FT: 1.58 and Vicuna: 1.10*) than the four ECS SOTA models and human responses. GPT's performance was good for the interpretation dimension (1.00). Interestingly, only Vicuna_FT yielded the best performance for the exploration dimension and not the other two LLMs.

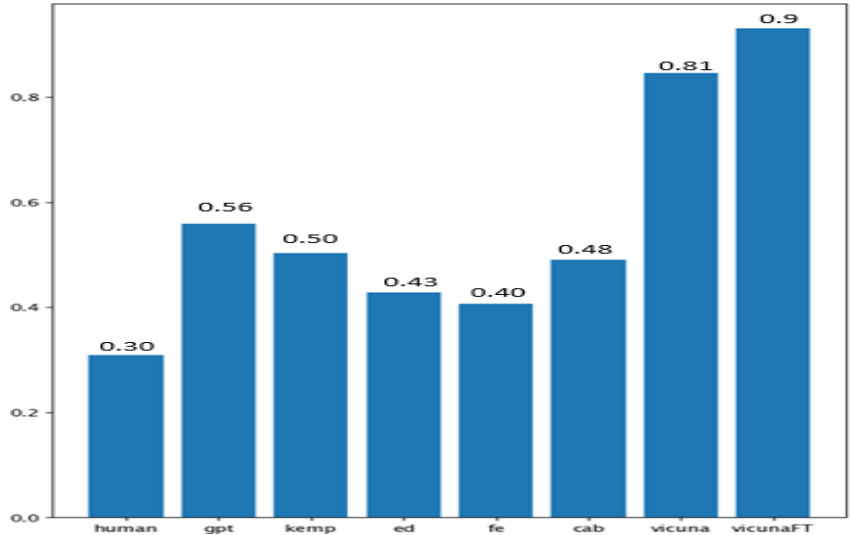

(a)

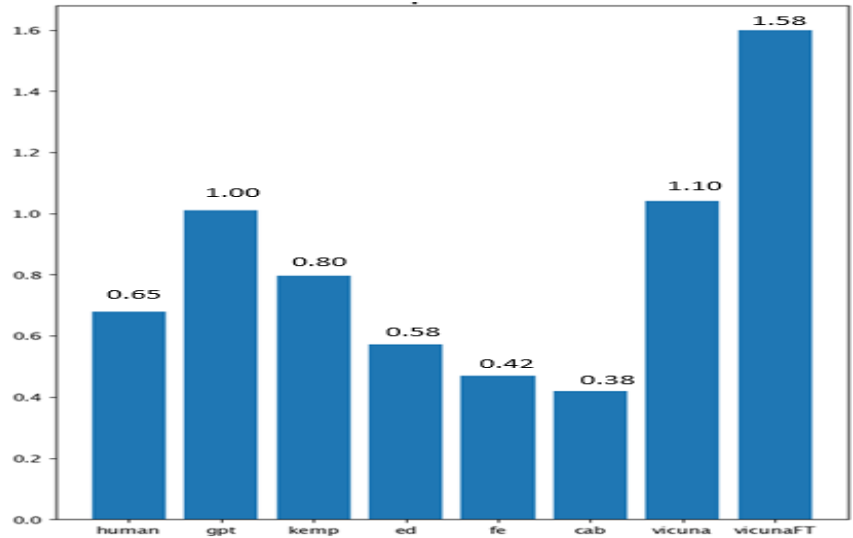

(b)

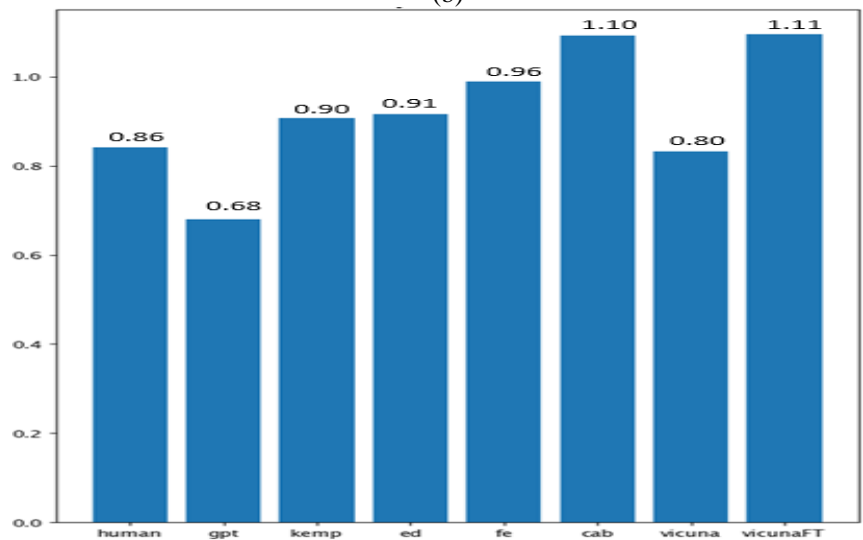

(c)

**Fig. 3.** Empathetic Responses Comparison with the three Empathy-related dimensions. (a) Emotional Reaction (b) Interpretation and (c) Exploration
*Note: In Y-axis, the average scores are plotted while the different models are part of the X-axis.*

## 2.2 Empathetic Behavior Evaluation

Beyond the structural level, there are intrinsic empathic behavior types identified in the empathy literature [13]. They are *mirroring, empathic concern, consolation, altru-* *istic helping,* and *perspective-taking.* The definitions of the behavior types are provided in Table 1. These behaviors provide a deeper understanding of an empathetic response. In addition to these behavior types, there is a need to determine the presence of *judgmental* and *apathic behaviors* in the response, as these qualities are essentially unempathetic. The measurement of these behavior types can be traditionally performed with two methods – algorithmic, and human perception questionnaires where the former involves using a series of computational steps for measurement while the latter involves the usage of survey questions and n-point Likert scales (for e.g., 5-point or 3-point) for each behavior types. Recently, a new method has emerged with the advent of LLMs. Since LLMs are inbuilt with reasonable NLU capabilities, researchers have utilized LLMs as evaluators by giving specific instructions. For instance, researchers have used GPT-4 to evaluate responses using different criteria [14]. On a related front, studies have reported successful usage of LLMs for data annotation tasks [15]. Hence, the LLM-based method of evaluation has surfaced as a viable alternative to the traditional methods.

**Table 1.** Empathetic Behavior Types

| Behavior Type | Definition |
|---|---|
| Mirroring | The process of automatically and unconsciously imitating the facial expressions, body language, and tone of voice of another person. This helps us to understand and share their emotional state [16]. |
| Empathic Concern | the feeling of sympathy and care for another person who is in need. It is an other-oriented emotion that is elicited by the perceived welfare of another person [17]. |
| Consolation | It is the act of providing comfort or support to someone who is experiencing distress [18]. |
| Altruistic Helping | It is the act of helping someone else, even when there is no personal gain or reward involved [18]. |
| Perspective Taking | It is a cognitive skill that involves putting oneself in another person's shoes and seeing the world from their perspective [19]. |

We used GPT3.5 as an evaluator for the empathetic behavior types (mirroring, empathic concern, consolation, altruistic helping, and perspective-taking). Eight responses (four state-of-the-art ECS models, three LLMs, and human responses) for the conversations from the ED dataset were compared through this evaluation. The following prompt was provided to GPT3.5 – "*You will be given a dialog history and responses from eight different assistants. Please identify whether the responses indicate the presence of 'empathetic concern' characteristic in them in relation to the user utterances in the dialog history. Please make sure to provide the output only as 'yes' or 'no' for each response".* This prompt was modified for each empathetic behavior type. Along with the five empathetic behavior categories, we instructed GPT3.5 to provide yes/no output for two more metrics – judgmental response and apathetic response.

The evaluation results are illustrated with column charts in Figure 4. The percentages in the chart indicate the proportion of responses for each model that received a 'yes' value from GPT3.5. Four patterns could be identified from these charts – (i)

GPT was the best model closely followed by Vicuna, (ii) Finetuning seems to have improved the Vicuna model for all the constructs at a substantial level with the highest improvement for mirroring (*from 78.52% to 94.08%*), (iii) the human responses were the next best after LLMs and (iv) the four ECS SOTA models were the worst performing models with much scope for improvement. With this evaluation component, we could demonstrate that LLMs show more promise than traditional ECS models for empathy generation. Among the LLMs, the finetuned Vicuna LLM had scores very close to GPT3.5.

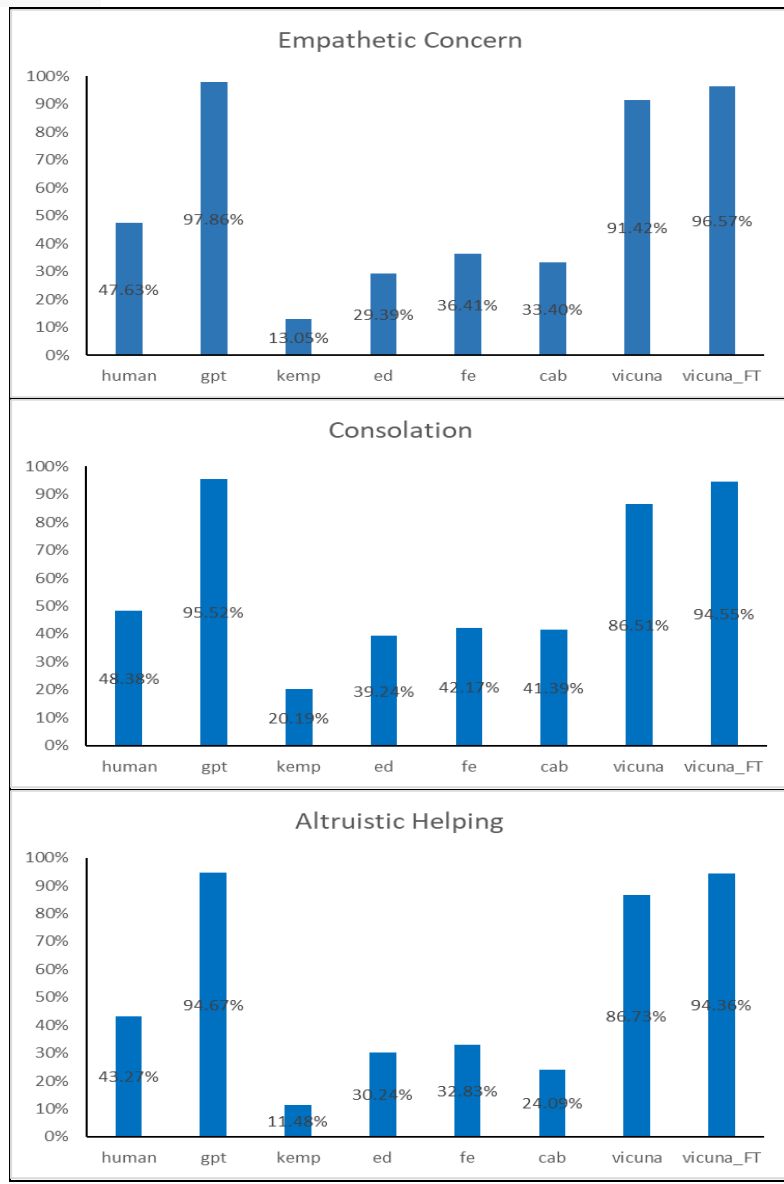

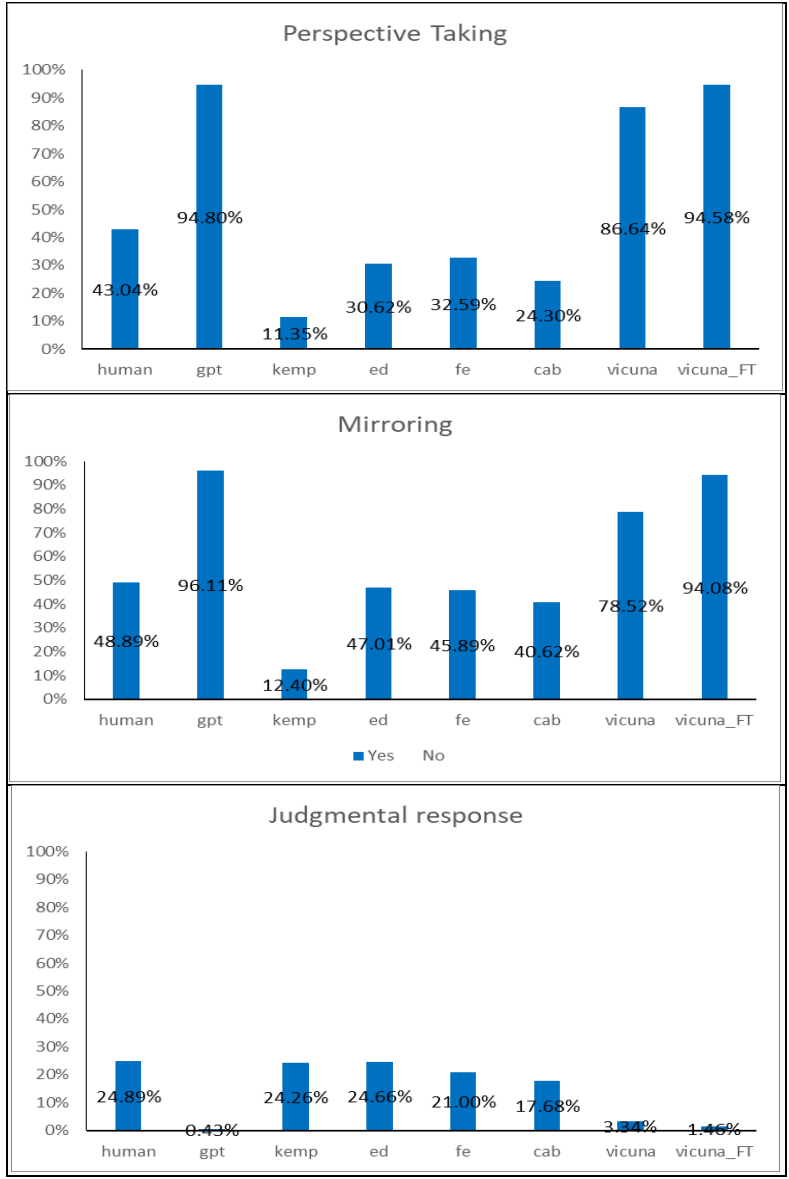

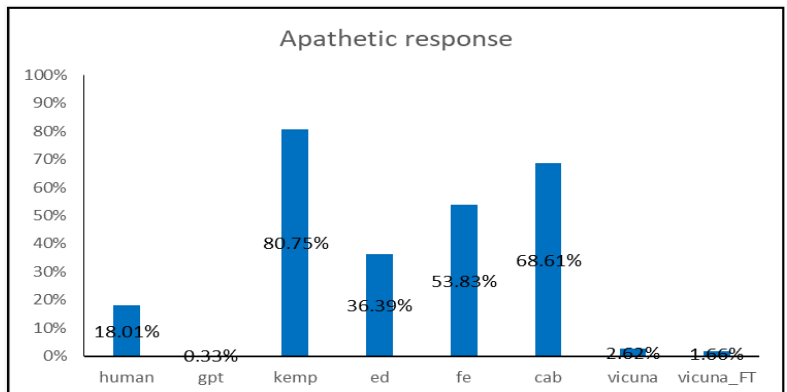

**Fig. 4.** Presence of Empathetic Behavior Types in the Different Models

### 2.3 Empathy Lexicon-based Evaluation

This lexicon approach is inspired by VADER [20] and CrystalFeel [21] for calculating the intensity scores for emotion and sentiment classes. The first step in this evaluation method is the creation of a lexicon with words and phrases representing different empathetic behavior types. Weights are assigned to the phrases for ranking purposes. The second step is to devise an algorithm to score the response based on the presence of words and phrases that match the lexicon entries. For creating the empathy lexicon, high quality responses from the ECS datasets could be considered as candidates for empathetic words and phrase extraction. This evaluation method eliminates the bias associated with both human and LLM evaluations but there is a caveat that imposters could game the lexicon by merely placing empathetic words in the response. Hence, the scoring algorithm needs to additionally consider the relevance of the response to the input utterance/dialog history before calculating the empathy score.

### 2.4 Offline Evaluation

If there is an apparent deficit of strong empathetic quality in the ECS datasets at an overall level, subsets could be created from the datasets by selecting the dialogues with high empathy scores for finetuning the ECS models or LLMs. In such a scenario, gold standard comparison using offline evaluation metrics is a standard choice for evaluating the ECS models. Offline evaluation is the most-used evaluation approach in ECS studies [4]. In offline evaluation, the ECS models are evaluated on their ability to reproduce the listener's responses to a speaker from the datasets. The metrics Perplexity, Bilingual Evaluation Understudy Score (BLEU) and Distinct-n (dist-n) are the most frequently used offline evaluation metrics. Earlier ECS studies [6, 7] provide guidelines on how to use these metrics to perform offline evaluation. This evaluation approach is utilized in existing ECS studies for benchmarking purposes.

In Table 2, we have listed the results from an offline evaluation experiment which was conducted with the same set of models with the test set of the ED dataset. Three evaluation metrics were used. Perplexity is used to evaluate how well a model can predict the next word or token in a conversation. A lower perplexity score is generally considered better, as it indicates that the model's predictions are more certain and closer to the true distribution of words or tokens in the given context. The BLEU metric primarily measures the overlap of n-grams between the generated text and reference (human-generated) text [22]. The score typically ranges from 0 to 1, with 1 indicating a perfect match between the generated and reference text. The 'Distinct-1' metric measures the diversity of unigrams in the responses generated model [23]. It assesses how many unique words are used in the model's responses. The value range is from 0 to 1 with 1 being highly unique and 0 being no unique words in the response.

The best (lowest) perplexity mean score among all the models is from the FE (8.322) followed by KEMP (9.219) and CAB (9.249). These three models are ECS models that are trained to simulate the conversations in the ED dataset. Similarly for the BLEU scores., the values for the ECS models are better than the LLMs. With the Distinct-1 metric, the human baselines and the corresponding ED model have the best values with 0.955 and 0.959 respectively followed by the ECS models and LLMs. These results indicate the ECS models have the best evaluation results in terms of simulating the conversations in the ED dataset. However, the calculation heuristics for these metrics do not involve measurement of any empathy component, hence we cannot confidently claim that the model responses will have the adequate quality of empathy embedded in them.

**Table 2.** Offline Evaluation Results

| Model | Perplexity Mean (SD) | BLEU Mean (SD) | Distinct-1 Mean (SD) |
|---|---|---|---|
| Human | - | - | 0.955 (0.063) |
| GPT | 42.568 (14.295 | 0.009 (0.013) | 0.811 (0.081) |
| KEMP | 9.219 (2.528) | 0.027 (0.051) | 0.947 (0.07) |
| ED | 10.077 (3.271) | 0.024 (0.044) | 0.959 (0.058) |
| FE | 8.322 (3.112) | 0.027 (0.067) | 0.93 (0.09) |
| CAB | 9.249 (2.719) | 0.027 (0.051) | 0.898 (0.13) |
| Vicuna | 41.595 (18.378) | 0.009 (0.018) | 0.702 (0.156) |
| Vicuna_FT | 39.432 (20.047) | 0.012 (0.028) | 0.768 (0.142) |

## 2.5 Human Evaluation

The final evaluation component is dedicated to human evaluation. This evaluation has immense practical implications when conducted in a rigorous manner. As per our review, earlier ECS studies have used human evaluation by recruiting crowd workers to evaluate entire dialogue history through three constructs (i) Empathy (ii) Relevance and (iii) Fluency [4]. The questions for these three constructs are *"Did the responses show understanding of the feelings of the person talking about their experience?"*, "*Did the responses seem appropriate to the conversation? Were they on-topic?*" and *"Could you understand the responses? Did the language seem accurate?"* respectively.

Instead of human subjects evaluating the whole dialogue, it is apt to evaluate each empathetic response. We propose pertinent questions that consider the different empathetic behavior types for each empathetic response. Secondly, we propose questions that are meant to evaluate the overall dialogue quality. The questions are listed in Table 3. We recommend the usage of a five-point Likert scale to measure the responses with the values *Strongly Disagree (1), Disagree (2), Neutral (3), Agree (4), Strongly Agree (5).*

**Table 2.** Questions for Human Evaluation

| Level | Question |
| --- | --- |
| Utterance-level | Is the supporter reacting emotionally to the seeker's utterance or trying to mirror the seeker's emotions? |
| | Is the supporter attempting to explore the seeker's situation further? |
| | Is the supporter attempting to understand or interpret the seeker's situation? |
| | Does the supporter's response appear judgemental? |
| | Does the supporter's response indicate lack of interest or concern towards the seeker? |
| | Does the supporter's response indicate genuine concern towards the seeker's situation? |
| | Is the supporter attempting to console the seeker? |
| | Is the supporter attempting to selflessly help the seeker without expecting anything in return? |
| Dialogue-level | The conversation is relevant to the emotion type - <emotion class> |
| | The supporter exhibits sufficient emotion towards the seeker's issues |
| | The supporter fully understood the seeker's issues |
| | The supporter put himself/herself in the seeker's position and addressed the seeker's issues |
| | The supporter provided a solution/intervention or suggestion to the seeker |
| | The emotional distress of the seeker improved towards the end of the conversation |

# 3 Conclusion

This paper contributes to the topics of strategy and methods for evaluating the empathic responding capability of ECS, LLMs, and dialogue/conversational systems in general. A multidimensional evaluation framework is proposed with three novel components to comprehensively measure empathy using a variety of metrics and measuring mechanisms, leveraging both human and system intelligence. For three of the five components in the framework, we conducted experiments with the conversations from the test set of the EmpatheticDialogues (ED) dataset. We compared the performance of four SOTA ECS models and three LLMs along with baseline conversations. The LLMs performed the best for the empathy-related components of the framework while the ECS models performed well for the traditional offline evaluation metrics. For future ECS studies, this framework serves as a comprehensive guide for evaluating the empathy related concepts in the conversations. As a part of the future work, we plan to devise an approach for building the empathy lexicon proposed in this paper and conduct the human evaluation of the different models.